\RequirePackage{fix-cm}
\documentclass[smallextended]{svjour3}       
\smartqed  
\usepackage{graphicx}
\usepackage{pdfpages}
\usepackage{epstopdf}
\usepackage{booktabs}
\usepackage{setspace}
\usepackage{diagbox}
\usepackage{tipa}
\usepackage{stmaryrd}
\usepackage{pifont}
\usepackage{amssymb}
\usepackage{amsmath}
\usepackage{cite}
\usepackage{amsfonts}
\usepackage{txfonts}
\usepackage{amsmath}
\usepackage{latexsym}
\usepackage{cases}
\usepackage{subfigure}
\usepackage{graphicx}
\usepackage{algorithm}
\usepackage{algorithmic}
\usepackage{bbding}
\usepackage{times}
\usepackage{array}
\usepackage{color}
\usepackage{mathrsfs}

\begin{document}

\title{Edge Data Based Trailer Inception Probabilistic Matrix Factorization for Context-Aware Movie Recommendation}

\author{Honglong Chen \and Zhe Li \and Zhu Wang \and Zhichen Ni \and Junjian Li \and Ge Xu \and Abdul Aziz \and Feng Xia}



\institute{H. Chen, Z. Li, Z. Wang, Z. Ni, and J. Li \at
              College of Control Science and Engineering, China University of Petroleum, Qingdao 266580, China. \\
               \email{chenhl@upc.edu.cn}
            \and
           H. Chen and G. Xu \at
             College of Computer and Control Engineering, Minjiang University, Fuzhou 350108, China \\
              \email{xuge@pku.edu.cn}
           \and
           A. Aziz \at
             School of Software, Dalian University of Technology, Dalian 116620, China. \\
            \and
           F. Xia \at
             School of Engineering, IT and Physical Sciences, Federation University Australia, Ballarat, VIC 3353, Australia. \\
              \email{f.xia@ieee.org}
}

\date{Received: date / Accepted: date}
\date{Corresponding author: Feng Xia. Email: f.xia@ieee.org.}

\maketitle

\begin{abstract}
 The rapid growth of edge data generated by mobile devices and applications deployed at the edge of the network has exacerbated the problem of information overload. As an effective way to alleviate information overload, recommender system can improve the quality of various services by adding application data generated by users on edge devices, such as visual and textual information, on the basis of sparse rating data. The visual information in the movie trailer is a significant part of the movie recommender system. However, due to the complexity of visual information extraction, data sparsity cannot be remarkably alleviated by merely using the rough visual features to improve the rating prediction accuracy. Fortunately, the convolutional neural network can be used to extract the visual features precisely. Therefore, the end-to-end neural image caption (NIC) model can be utilized to obtain the textual information describing the visual features of movie trailers. This paper proposes a trailer inception probabilistic matrix factorization model called Ti-PMF, which combines NIC, recurrent convolutional neural network, and probabilistic matrix factorization models as the rating prediction model. We implement the proposed Ti-PMF model with extensive experiments on three real-world datasets to validate its effectiveness. The experimental results illustrate that the proposed Ti-PMF outperforms the existing ones.
\end{abstract}

\section{Introduction}
\label{sec1}
In recent years, with the rapid development of Internet of Things (IoTs) \cite{Aixin:2021,Lin:2021,Chen:2022}, the number of mobile devices and applications deployed at the edge of the network to provide users with various services has increased significantly \cite{He:2020Diversified,Cui:2021Inter,Zhou:2020DRAIM,Zhou:2021Incentive,Liu:2021FedCPF}. However, although they make full use of the computing resources of edge servers, which makes a great contribution to reducing network latency \cite{Lai:2021Dynamic}, they also generate a large amount of edge data, which aggravates the problem of information overload faced by the contemporary era, thus affecting the quality of various services and the user satisfaction \cite{EdgeRec}. Recommender systems can extract user preferences by using various user behaviors and application data generated on edge devices, so as to generate corresponding recommendations \cite{Altulyan:2020Rec}. Unfortunately, the rapid growth of the number of users and the amount of related edge data makes data sparsity a challenging issue of the recommender systems \cite{G.Penha,Chen:2021,Li:2021}, which severely deteriorates the recommendation performance. As an effective way, the contextual feature \cite{S.Y.Chou} can be utilized to customize its recommendation by adding additional information, such as visual and textual features, to alleviate the sparseness \cite{wang2019tcss,GSIRec,VRConvMF2021,Wan2022Edge}. The original context-aware recommendation method is to get implicit feedbacks from users (such as the time spent on each page and click-through rate of each item) to infer whether the user prefers a certain item. In recent years, the deep learning-based algorithms have been well studied, for example, convolutional neural network (CNN) \cite{R.Gao} and recurrent neural network (RNN) \cite{Q.Cui} have contributed greatly to extracting the visual features and textual features, respectively. Additionally, the movie trailer often contains a lot of important information of the whole movie. Therefore, in movie context-aware recommender system, extracting the features of movie trailers through deep learning-based algorithms would bring valuable and reliable additional information.

Most of the previous context-aware movie recommendation methods just used the static contextual features (such as the user attribute and movie attribute) to improve the recommendation performance. However, with the recent advances of deep learning algorithms, it is not hard to capture the deep features of images and videos by using deformations of CNN, such as AlexNet \cite{J.Zhang}, GoogLeNet \cite{C.Szegedy} and so on. Accordingly, after extracting the multiple features, the feature combination should be considered in the movie recommendation \cite{Chen:2021,xia2013recAccess}. Therefore, an end-to-end network neural image caption (NIC) generator combining the visual feature extraction network and textual generation network is proposed \cite{Vinyals:2015}, which converts the movie trailer information into corresponding description texts. In the NIC model, the inputs are images, and the outputs are sentences, which are translated based on the visual features of the input images. The NIC method can speed up the trailer processing, making it feasible to utilize the visual features in a context-aware recommendation.

This paper focuses on utilizing the trailer information of movies in the context-aware recommendation to promote the performance of rating prediction. In order to avoid the coupling among different still frames, which may lead to a high similarity of the final visual features, we take 20 still frames evenly from one video track as the input of the NIC model. These still frames are input into the NIC model to generate the corresponding descriptive textual information, based on which the most accurate description texts can be extracted as the contextual text information. Note that the length of the descriptive texts of still frames extracted from movie trailers will be shorter than that of the users' review texts \cite{Z.Li}. Moreover, it can not determine whether the user prefers the movie or not only with some review sentences, which results in low quality of textual features for the review texts. While for the descriptive text, the sentence is specifically used to describe the movie trailers, which is more concise and more valuable. Therefore, the image information description is more effective than the feature information of the review texts for the recommender systems. In this paper, we propose a trailer inception probabilistic matrix factorization model called Ti-PMF, which integrates the movie visual information to alleviate the sparsity of rating data and promote the performance of rating prediction. In the experiments, we utilize the advanced visual feature extraction network VGG and GoogLeNet respectively to evaluate the text conversion performance, and finally, adopt the texts generated by NIC to get the corresponding root mean squared error. The experimental results illustrate that the proposed Ti-PMF model significantly outperforms the existing schemes.

{\bf The main contributions of this paper} are as follows:

\begin{itemize}
\item We propose a trailer inception probabilistic matrix factorization model called Ti-PMF, which integrates the movie visual information to alleviate the sparsity of rating data and promote the performance of rating prediction.

\item We utilize the NIC model to automatically convert the video information of the movie trailers into the corresponding descriptive textual information and then embed the textual information into our recommendation model seamlessly.

\item The training time of the proposed Ti-PMF method can be significantly shortened with a higher rating prediction accuracy.

\item We implement the proposed Ti-PMF model with extensive experiments on three real-world datasets to validate its effectiveness.

\end{itemize}

The rest of the paper is organized as follows. Sect.~\ref{sec2} introduces the related work of context-aware recommendation and feature engineering in deep learning. Sect.~\ref{sec3} briefly reviews preliminaries on the probabilistic matrix factorization, the neural image caption generator, and textual feature extraction of recurrent convolutional neural network. Sect.~\ref{sec4} concretely presents our proposed model Ti-PMF. Sect.~\ref{sec5} shows the experiments about the proposed model and discusses the results of our model. Sect.~\ref{sec6} summarizes our work.
\section{Related Work}
\label{sec2}
\subsection{Context-aware Recommendations}
With the advent research on context-aware processing and becoming a hot-spot research topic in the field of recommendation, it is considered that when more contextual information is provided, better recommendation accuracy can be achieved \cite{Asabere2014thms,liu2017systems}. The contextual information used for the context-aware recommendations includes time, location, entity or event. The context-aware recommender system modifies the existing model to a scene in the specific dimension, realizing the context directly embedded in the recommendation process. The method provides a flexible and generalizable context-aware recommendation, which overcomes the obstacles of two-dimensional algorithms \cite{L.B}. Specifically, the context-aware recommendation method can easily embed the contextual information. This method is applied to each item to make a precise recommendation for a given user $u$ with context $t$, and then the top-k item recommendation is accomplished. The methods for calculating neighborhood are existing, and collaborative filtering and content-based methods are very commonly utilized in recommender systems \cite{L.Zhao,Zhou2015}. The latent factor model (LFM) recommends items with similar item features to target users based on the element features in the users' contexts. Another segment of the application of LFM is the factorization machine (FM) \cite{W.Pan}, the ratings are modeled as linear combinations of the interactions between input variables of the model. In addition, machine learning algorithms are utilized in content-based recommender systems to extract attributes associated with users, items, and contexts \cite{Q.Zhu}. Moreover, in the context-aware recommender system, selecting appropriate attributes is also an important process. Common contextual information is utilized in recommender systems including: time, location, and social information \cite{xia2014exploiting,xia2014community}. Zarzour \textit{et al}. \cite{H.Z} proposed the conception of multidimensional attributes, which uses dimensionality reduction and clustering techniques to integrate multidimensional attributes and reduce their dimensionality, thereby obtaining attributes with higher accuracy. The LSIC model proposed in \cite{W.Zhao2019} explores context-aware information (movie posters) and uses GAN framework to leverage the matrix factorization (MF) and RNN approaches for top-N recommendation to further improve the performance of movie recommendation. Recently, a deep learning recommendation framework that incorporates contextual information into neural collaborative filtering recommendation approaches is proposed in \cite{Unger2020}, which models contextual information in various ways for multiple purposes, such as rating prediction, generating top-k recommendations, and classification of users' feedback. Chen \textit{et al}. \cite{L.Chen} proposed that the existing methods suffer from context redundancy, and proposed a context-aware recommendation method based on embedded feature selection, which eliminates context redundancy by generating a minimum subset of all contextual information, and allocates weight to each context appropriately to achieve performance improvements.

\subsection{Feature Engineering in Deep Learning}
 Feature engineering in deep learning embeds additional information into context-aware recommender system, which helps to alleviate the sparsity problem of rating data \cite{L.Chen}. The data presented to the algorithm by feature engineering has the relevant structure or attributes of the basic data of the corresponding task \cite{J.Liu,XF.Ding}. Since there are many types of existing attribute features, the single feature \cite{A.Q.Macedo} and the multi-feature fusion context-aware recommender system can be used \cite{K.Ram}. Besides, feature fusion between texts is often used in text classification tasks and multiple feature weighting \cite{Charu.C}. By preprocessing the data structure through feature engineering, the algorithm can reduce noise interference and find data trends. Specifically, in the preprocessing method based on dimensionality reduction, an item can be divided into several fictitious items by using several corresponding contexts in order to determine its attribute features \cite{H.Z}.~Zhang~\textit{et al}. proposed a method of using context to establish user portrait features for recommendation \cite{Z.H}. In the movie context-aware recommender system, multiple attribute features extracted from movie descriptive texts and user review texts are utilized to obtain the personalized recommendations \cite{Y.Kim}. To select high-quality feature representations, Goldberg \textit{et al}. discussed feature selection metrics for data classification \cite{M.G}. Dense matrix data features utilize cross-feature and feature fusion technology to project attributes onto the fixed dimensional data feature spaces. Therefore, in the field of feature engineering, fewer data dimensions can be utilized to express more attributes \cite{T.Mikolov,J.Pennington}.
 \section{Preliminary}
 \label{sec3}

\subsection{Probabilistic Matrix Factorization}
Probabilistic Matrix Factorization (PMF) \cite{Salakhutdinov:2008} can obtain a relatively accurate prediction based on a few specific scores in the rating matrix. PMF aims to improve the rating prediction accuracy of the conventional matrix factorization by using the probabilistic method. Specifically, it is supposed that there are $M$ movies and $N$ users. The element $R_{ij}$ in rating matrix $R\in\mathbb{R}^{N\times M}$ represents the rating of user $i$ on movie $j$. The number of latent features is expressed as $D$, user matrix $U\in\mathbb{R}^{D \times N}$ and movie matrix $V\in\mathbb{R}^{D \times M}$ are both latent feature matrices, and their column vectors $U_i$ and $V_j$ represent the user-specific and the movie-specific latent feature vectors, respectively. Then PMF is based on the following two assumptions: 1) the observed errors follow the Gaussian distribution, 2) the user matrix $U$ and movie matrix $V$ follow the Gaussian distribution. The conditional distribution over the observed ratings based on the above two assumptions is:
\begin{equation}
p\Big(R|U,V,{\sigma ^2}\Big)=\mathop \prod \limits_{i = 1}^N \mathop \prod \limits_{j = 1}^M {\bigg[ N\Big({R_{ij}}|U_i^T{V_j},{\sigma ^2}\Big)\bigg]^{{I_{ij}}}},\label{eq1}
\end{equation}
where $N(x\mid\mu,\sigma^2)$ is the probability density function, which conforms to the Gaussian distribution with mean $\mu$ and variance $\sigma^2$. $I_{ij}$ is an indicator function, if user $i$ rated movie $j$, the function is equal to 1, otherwise it is equal to 0. The zero-mean spherical Gaussian priors are considered on the user and movie feature vectors and can be formulated as:
\begin{equation}
p\Big(U|\sigma _U^2\Big)=\mathop \prod \limits_{i = 1}^N N\Big({U_i}|0,\sigma _U^2I\Big), \label{eq2}
\end{equation}
\begin{equation}
p\Big(V|\sigma _V^2\Big)=\mathop \prod \limits_{j = 1}^M N\Big({V_j}|0,\sigma _V^2I\Big). \label{eq3}
\end{equation}
Note that $I$ in the above equation is not an indicator function, it represents a diagonal matrix. The $L2$ regularization term is applied to avoid over-fitting, and the loss function can be formulated as:
\begin{small}
\begin{equation}
\mathcal{L}(U,V)=\frac{1}{2}\sum\limits_{i=1}^N\sum\limits_{j=1}^MI_{ij}\Big(R_{ij}-U_i^T{V_j}\Big)^2+\frac{\lambda_{U}}{2}\sum_{i=1}^M\Big|\Big|U_{i}\Big|\Big|_F^2+\frac{\lambda_{V}}{2}\sum_{j=1}^M\Big|\Big|V_{j}\Big|\Big|_F^2,\label{eq4}
\end{equation}
\end{small}where ${\lambda _U} = \sigma^2 / \sigma_U^2,$ ${\lambda _V} = \sigma^2 / \sigma_V^2,$ and $\parallel\cdot\parallel^2_F$ denotes the frobenius norm.
\subsection{Neural Image Caption Generator}
\begin{figure}
  \centering\includegraphics[width=3.6in,height=2.3in]{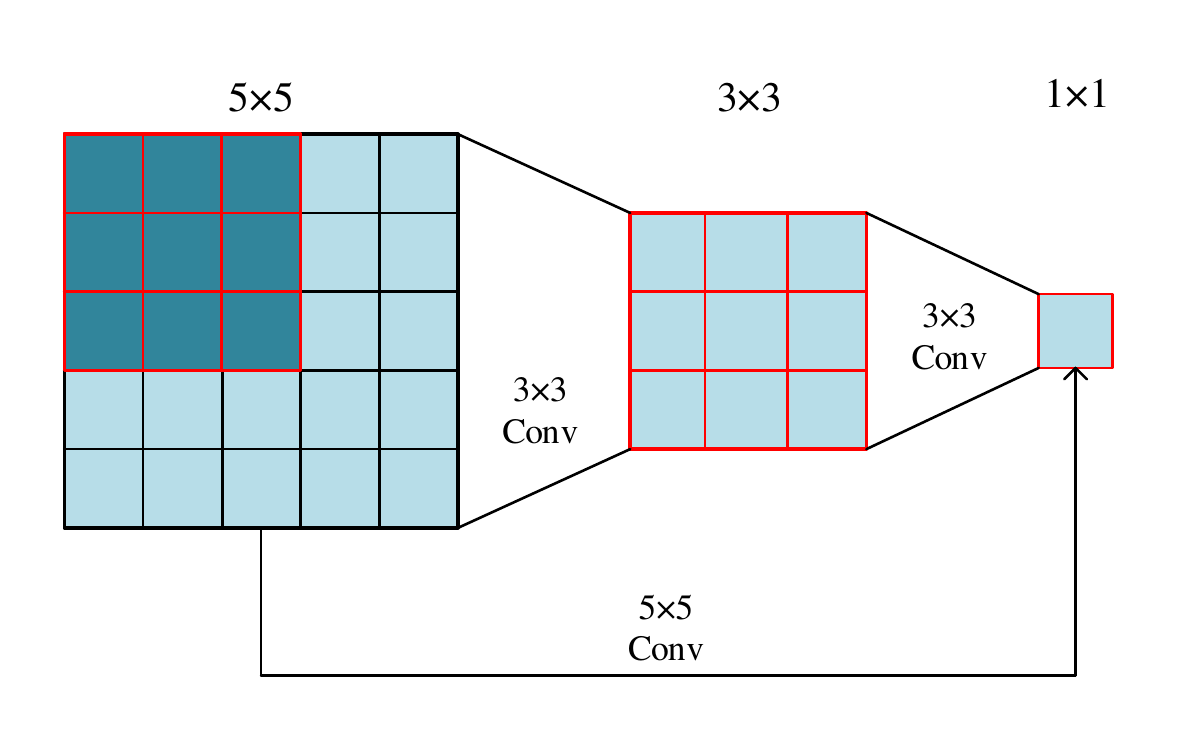}
  \caption{A 5*5 filter is replaced with double 3*3 filters (stride = 1).}
  \label{figure1}
\end{figure}
\begin{figure}
  \centering\includegraphics[width=3.6in,height=1.82in]{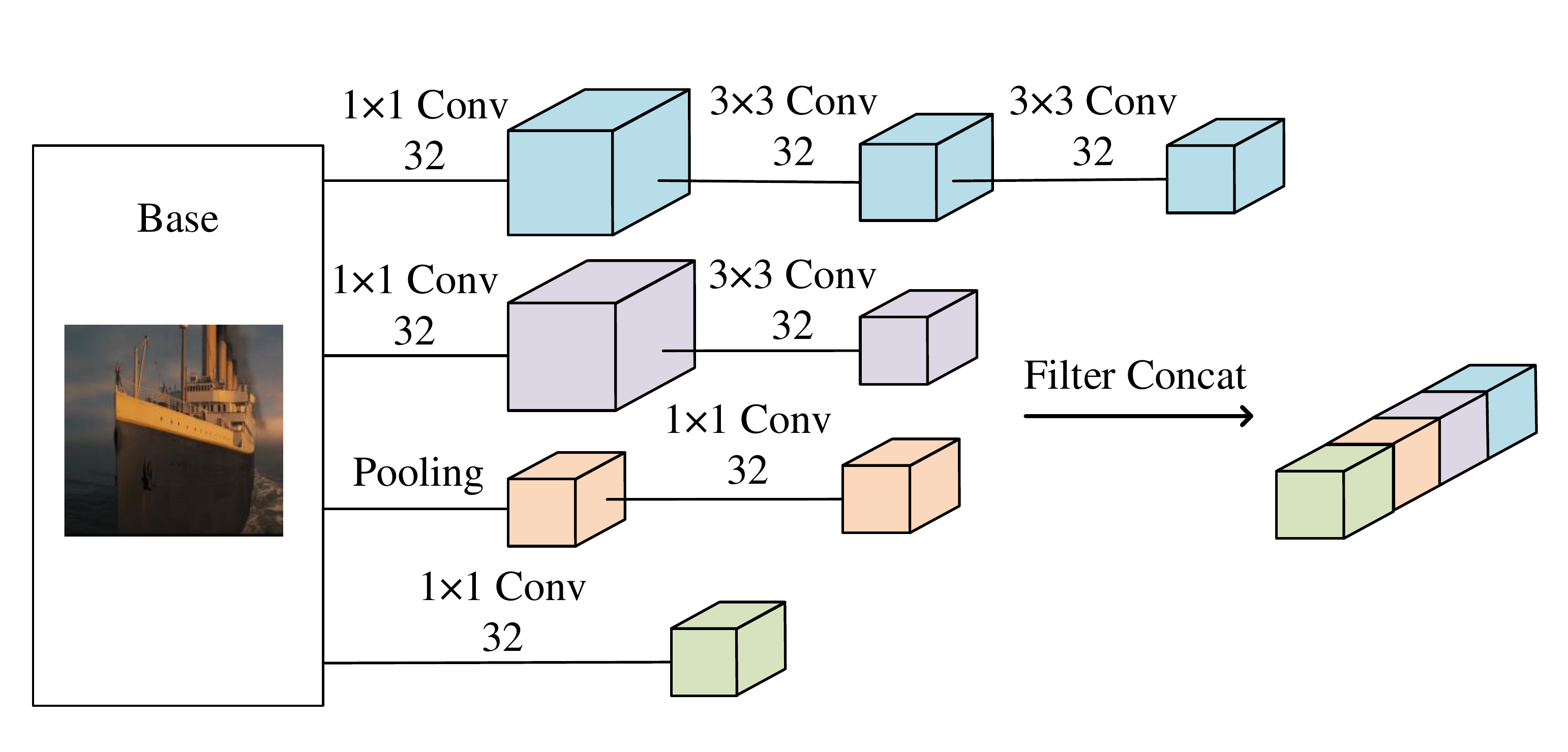}
  \caption{The architecture of basic inception model with multiple convolutional kernels (inception-a).}
  \label{figure2}
\end{figure}

Neural image caption (NIC) generator is an end-to-end network with images as the input and text sequences as the output. Specifically, NIC integrates GoogLeNet for extracting the images' visual features and RNN for converting the visual features into sequential texts.

\subsubsection{GoogLeNet}

\begin{figure}
  \centering\includegraphics[width=3.3in,height=2.0in]{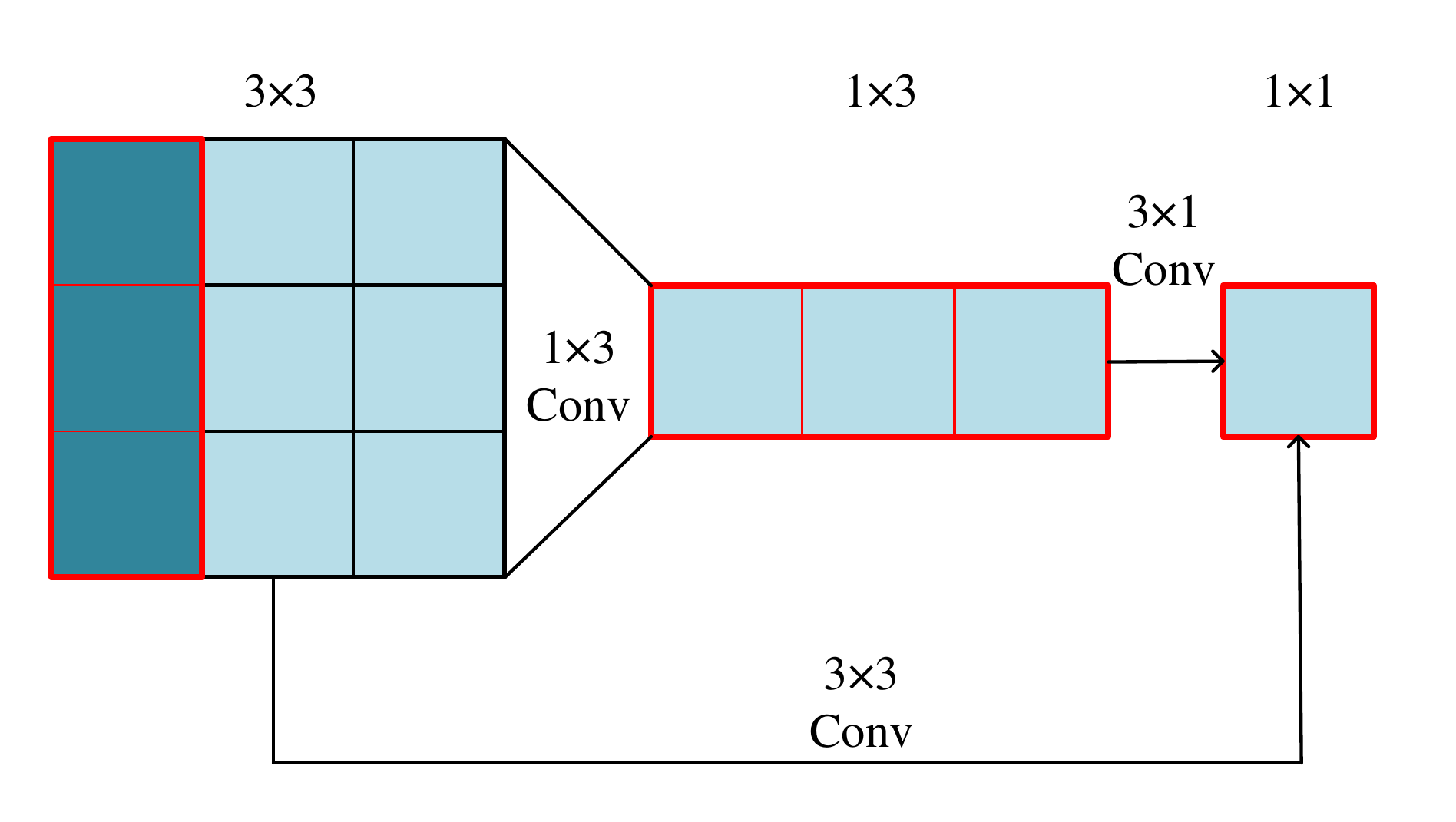}
  \caption{A 3*3 filter is factorized by a 1*3 filter and a 3*1 filter (stride = 1).}
  \label{figure111}
\end{figure}

\begin{figure}[htbp]

\centering
\subfigure[Inception-b]{
\begin{minipage}[t]{0.5\linewidth}
\centering
\includegraphics[width=1.4in]{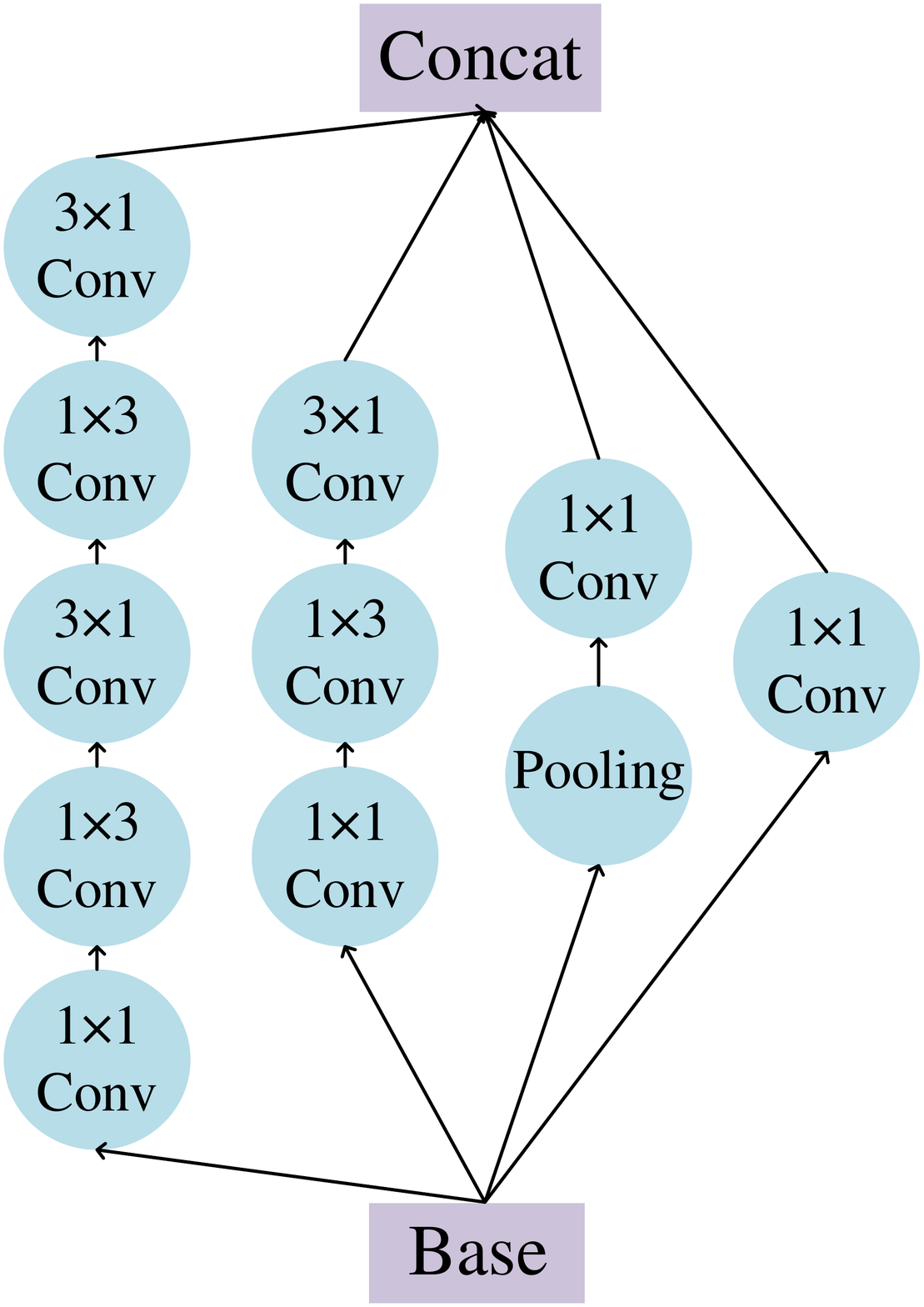}
\label{figure3}
\end{minipage}%
}%
\subfigure[Inception-c]{
\begin{minipage}[t]{0.5\linewidth}
\centering
\includegraphics[width=1.7in]{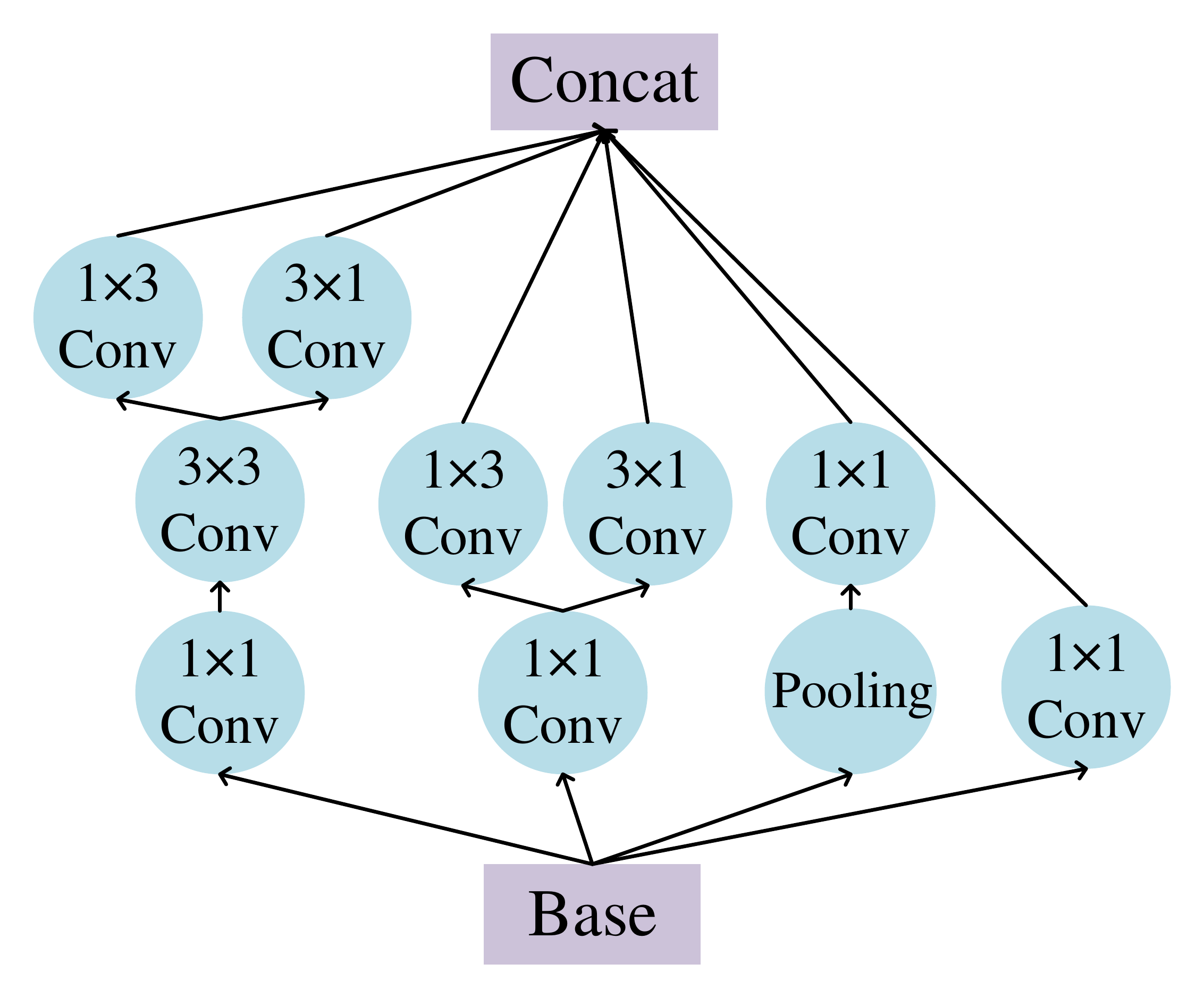}
\label{figure4}
\end{minipage}%
}%
\centering
\caption{The architectures of the Inception-b and Inception-c.}
\end{figure}

Before GoogLeNet, a deep learning structure, was proposed\cite{C.Szegedy1}, AlexNet, VGG and other structures used the method of increasing the depth of the network to achieve better training results, which would bring negative effects such as over-fitting, gradient disappearance, gradient explosion and so on. Inception is the basic component of the GoogLeNet network. The introduction of inception can make effective use of computing resources and obtain more features under the same amount of computation, which can improve the training effect. Specifically, various sizes of convolutional kernels ($1\times1$, $3\times3$, $5\times5$) are utilized, which can perceive different perceptive fields and obtain more comprehensive and richer visual feature information. Since the receptive field of the $5\times5$ convolutional kernel is the same as the receptive field of two $3\times3$ convolutional kernels, the number of training parameters, i.e., $18$, will be smaller than that of previous training parameters, i.e., $25$. Therefore, one $5\times5$ convolutional kernel can be replaced by two $3\times3$ convolutional kernels\cite{S.Ioffe}, as shown in Fig.~\ref{figure1}. And Fig.~\ref{figure2} shows the structure of inception\cite{C.Szegedy}. Besides, $1\times1$ convolutional kernel and pooling operation are used to compute reductions before the expensive $3\times3$ and $5\times5$ kernels. In the GoogLeNet model, $1\times1$ convolutional kernel is mainly utilized for dimensionality reduction for image data. Although the large convolutional kernel is very helpful for extracting visual features, it will cause a parameter explosion in the deep neural network. Therefore, Szegedy~\textit{et al}. decomposed the large convolutional kernel asymmetrically to reduce the number of parameters\cite{C.Szegedy}. Asymmetrical convolutional structure splitting is better than symmetrical convolutional structure splitting in processing more and richer spatial features and increasing feature diversity when reducing the amount of calculation. Specifically, an $n\times n$ convolutional kernel can be replaced by two convolutional kernels, a $1\times n$ kernel followed by an $n\times1$ one. In addition, the computational cost saved by the replacement can increase significantly with the increase of $n$. Finally, the CNN structure is equivalently replaced by the mini factorization form. The specific value of $n$ is determined according to the size of the input images. For instance, using a $3\times1$ convolutional kernel followed by a $1\times3$ one is equivalent to sliding one layer network with the same receptive field as in a $3\times3$ convolutional kernel. Accordingly, the number of parameters in training will be reduced from $3\times3=9$ to $3+3=6$, which is shown in Fig~\ref{figure111}. The second and third forms of inception are to factorize part of big convolutional kernels ($3\times3$)\cite{C.Szegedy}, as shown in Fig.~\ref{figure3} and Fig.~\ref{figure4}. Finally, the above three inception blocks (Fig.~\ref{figure2},~Fig.~\ref{figure3},~Fig.~\ref{figure4}) are combined to the final visual neural network. The network structures of VGG16\cite{K.Simonyan}, VGG19\cite{K.Simonyan} and GoogLeNet\cite{C.Szegedy} are shown in TABLE~\ref{table4}.

\begin{table}
\caption{The network structures of VGG16, VGG19 and GoogLeNet.}\label{table4}
\begin{spacing}{1.2}
\centering
\begin{tabular}{|c|c|c|}
\hline
\multicolumn{3}{|c|}{Input}\\
\hline
VGG16&VGG19&GoogLeNet\\
\hline
$2\times$conv3-64&$2\times$conv3-64&conv3-32\\
\hline
pool2-64&pool2-64&conv3-32\\
\hline
$2\times$conv3-128&$2\times$conv3-128&pool3-64\\
\hline
pool2-128&pool2-128&conv3-64\\
\hline
$3\times$conv3-256&$4\times$conv3-256&conv3-80\\
\hline
pool2-256&pool2-256&conv3-192\\
\hline
$3\times$conv3-512&$4\times$conv3-512&$3\times$inception-a\\
\hline
pool2-512&pool2-512&$5\times$inception-b\\
\hline
$3\times$conv3-512&$4\times$conv3-512&$2\times$inception-c\\
\hline
pool2-512&pool2-512&pool8-2048\\
\hline
$2\times$fc1-4096&$2\times$fc1-4096&fc1-2048\\
\hline
fc1-1000&fc1-1000&softmax1-1000\\
\hline
\multicolumn{3}{|c|}{Output}\\
\hline
\end{tabular}
\end{spacing}
\end{table}

\subsubsection{Long-Short Term Memory}

In order to effectively process the long sequences and solve the problem of gradient disappearance, long-short term memory (LSTM) is proposed on the basis of traditional RNN. When LSTM is utilized to process the information of each neuron in sequences, the true meaning of the current word in the sequence is inferred by the understanding of a previously seen word. The memory cell $c$ is the core of the LSTM model, which encodes the information of the observed inputs at every time step. The cell's behavior is controlled by three different gates: input gate, output gate, and forget gate. When the value of forget gate is set as 1, the information in LSTM will be maintained, and 0 means that the information will be forgotten. In particular, the three gates are used to control whether to forget the current cell value (forget gate $f$), whether to read the input (input gate $i$), and whether to output the new cell value (output gate $o$). The definitions of the gates, cell update, and output are as follows:
\begin{equation}
i_t=\sigma(W_{ix}x_t+W_{im}m_{t-1}),\label{eq4}
\end{equation}
\begin{equation}
f_t=\sigma(W_{fx}x_t+W_{fm}m_{t-1}),\label{eq5}
\end{equation}
\begin{equation}
o_t=\sigma(W_{ox}x_t+W_{om} m_{t-1}),\label{eq6}
\end{equation}
\begin{equation}
c_t=f_t\odot c_{t-1} + i_t \odot h (W_{cx}x_t+W_{cm}m_{t-1}),\label{eq7}
\end{equation}
\begin{equation}
m_t = o_t \odot c_t,\label{eq8}
\end{equation}
\begin{equation}
p_{t+1} = Softmax(m_t),\label{eq9}
\end{equation}
where $\odot$ represents the product with a gate value. The nonlinearities are sigmoid $\sigma(\cdot)$ and hyperbolic tangent $h(\cdot)$. The matrices $W_{ix}$, $W_{im}$, $W_{fx}$, $W_{fm}$, $W_{ox}$, $W_{om}$, $W_{cx}$, and $W_{cm}$, are the trained parameters. In Eq.~(\ref{eq9}), $m_t$ is fed to the softmax function, resulting in a probability distribution $p_t$ over all words.

\subsection{Textual Feature Extraction in RCNN}

\subsubsection{Recurrent Structure in Convolutional Layer}

\begin{figure}
  \centering\includegraphics[width=3.7in,height=2.32in]{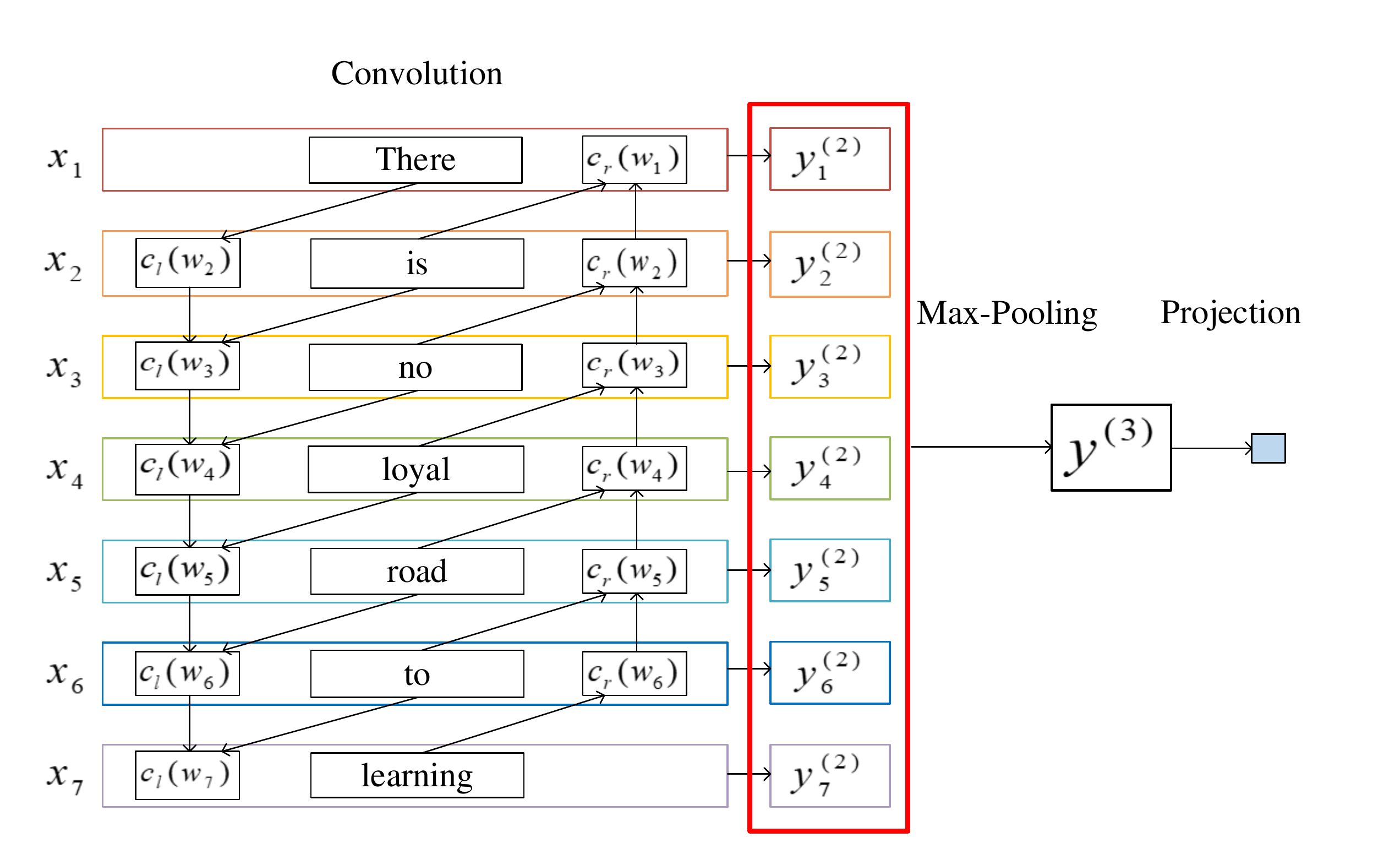}
  \caption{RCNN model used in this paper.}
  \label{figure7}
\end{figure}

Recurrent convolutional neural network (RCNN) \cite{S.Lai} model embeds the recurrent structure into the convolutional layer. On the one hand, CNN is utilized to extract the textual features. On the other hand, it can also use RNN structure to memorize the full-textual information. Meanwhile, the recurrent structure can obtain the contextual information as much as possible, which means that less noise may be introduced than the window-based neural networks. The features extracted by RCNN are used as a part of the mean of Gaussian distribution in the item latent models. Specifically, in the RCNN model, a word is combined with both of the left and right contexts to represent itself in the word representation. Consequently, a word representation of RCNN contains much richer information with its associated contextual information than that of CNN. The specific context expressions are as follows:
\begin{equation}
c_l(w_i)= ReLU\bigg(W^{(l)}c_l\Big(w_{i-1}\Big)+W^{(sl)}e\Big(w_{i-1}\Big)\bigg), \label{eq10}
\end{equation}
\begin{equation}
c_r(w_i) = ReLU\bigg(W^{(r)}c_r\Big(w_{i+1}\Big)+W^{(sr)}e\Big(w_{i+1}\Big)\bigg), \label{eq11}
\end{equation}
where $c_l(w_i)$ and $c_r(w_i)$ represent the left and the right contexts of the word $w_i$ respectively, $e\big(w_i\big)$ represents the word representation of word $w_i$, $W^{(sl)}$ and $W^{(sr)}$ represent the matrices, which are used to connect the semantics of the current word with the left and right adjacent words respectively. Furthermore, $W^{(l)}$ and $W^{(r)}$ represent the matrices, which combine all of the left and right context hidden layers, respectively. Then the context information and the word representation are cascaded as the whole word embedding model. Specifically, the word representation $x_i$ of word $w_i$ with its context information can be expressed as follows:
\begin{equation}
x_i = \bigg[c_l\Big(w_i\Big), e\Big(w_i\Big), c_r\Big(w_i\Big)\bigg].  \label{eq12}
\end{equation}
Note that different context window sizes can be utilized to capture different contextual information so as to investigate the performance more comprehensively. For instance, the word representation of $w_i$ is represented by $\Big[x\big(w_{i-1}\big); x\big(w_i\big); x\big(w_{i+1}\big)\Big]$ when the context window size is set to 3. Furthermore, an activation function, i.e., $\tanh$, is applied to transform $x_i$ into $y_i^{(2)}$ as follows:
\begin{equation}
y_i^{(2)} = \tanh\Big(W^{(2)}x_i + b^{(2)}\Big). \label{eq13}
\end{equation}
The RCNN model used in this paper is shown in Fig.~\ref{figure7}.

\subsubsection{Recurrent convolutional matrix factorization}
\label{sec3.2}
The generated texts are input into the recurrent convolutional matrix factorization (RConvMF) recommender system, which combines RCNN with PMF. Thereby, the movie latent model with visual trailer features is obtained by the following equations:
\begin{equation}
V_j = rcnn(W, X_j) + \varepsilon_j,  \label{eq13}
\end{equation}
\begin{equation}
\varepsilon_j \sim N(0,\sigma_V^2 I),  \label{eq14}
\end{equation}
where $X_j$ represents the descriptive texts converted from the visual information extracted from the movie trailers. In Eq.~(\ref{eq13}), for each weight $w_k$ in $W$, the zero-mean spherical Gaussian prior is shown as follows:
\begin{equation}
p(W|\sigma_W^2) = \prod\limits_k N(w_k|0, \sigma^2_W). \label{eq15}
\end{equation}
Accordingly, the conditional distribution over item latent models is given by:
 \begin{equation}
p(V|W,X,\sigma_V^2) = \prod\limits_j^M N(V_j|rcnn(W,X_j), \sigma^2_V I), \label{eq16}
\end{equation}
where $X$ is the set of descriptive documents of items obtained through the NIC model from the movie trailers. A document latent vector obtained from the RCNN model is taken as the mean of Gaussian distribution, and the Gaussian noise of the item is taken as the variance of the gaussian distribution. In this way, the NIC, RCNN, and PMF model can be connected seamlessly.
\begin{figure*}
  \centering\includegraphics[width=4.5in]{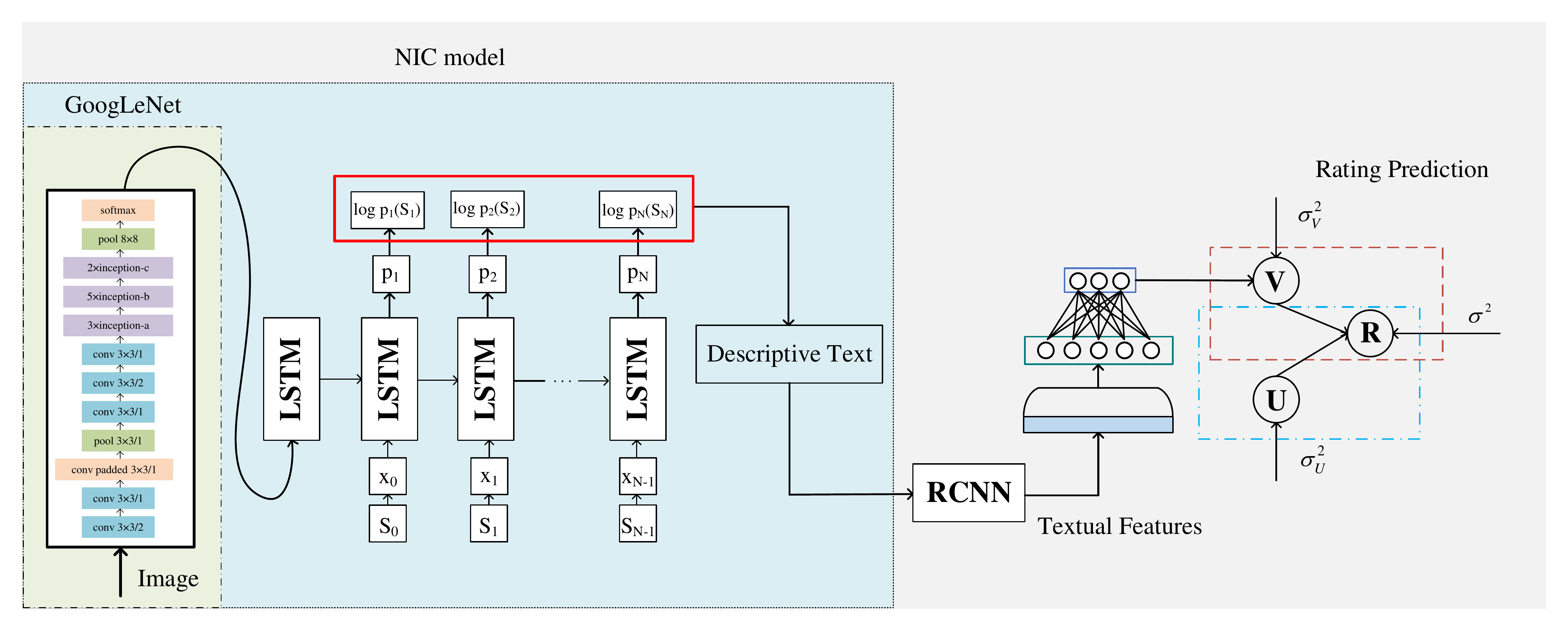}
  \caption{The architecture of the proposed Ti-PMF model.}
  \label{figure8}
\end{figure*}

\subsubsection{Optimization Methodology}

To optimize the variables such as the weights and bias of RCNN, the maximum a posteriori (MAP) estimation is utilized, which can be expressed as follows.
{\small\begin{equation}
\begin{split}
&\max\limits_{U,V,W}p\bigg(U,V,W\Big|R,X,Y,\sigma^2,\sigma_U^2,\sigma_V^2,\sigma_W^2\bigg)\\
&=\max\limits_{U,V,W}\bigg[p\Big(R\Big|U,V,\sigma^2\Big)p\Big(U\Big|\sigma_U^2\Big)p\Big(V\Big|W,X,Y,\sigma_V^2\Big)p\Big(W\Big|\sigma_{W}^2\Big)\bigg].\\\label{eq17}
\end{split}
\end{equation}}

Coordinate descent optimization method is adopted in training. It optimizes a latent variable iteratively while fixing the other ones. The optimal solution of $U_i$ and $V_j$ can be obtained as follows:
\begin{equation}
{U_i} \leftarrow {\Big(V{I_i}{V^T} + {\lambda _U}{I_K}\Big)^{ - 1}}V{R_i},\label{eq18}
\end{equation}
\begin{equation}
{V_j} \leftarrow {\Big(U{I_j}{U^T} + {\lambda _V}{I_K}\Big)^{ - 1}}\Big(U{R_j} + {\lambda_V}{\mu_j}\Big),\label{eq19}
\end{equation}
where $I_{i}$ is a diagonal matrix with $I_{ij}$, $j=1,\cdots,n$ as its diagonal elements and $R_{i}$ is a vector for user $i$ with $\big(r_{ij}\big)_{j=1}^n$. The back-propagation algorithm is applied to optimize $W$. In the whole optimization process, the unobserved rating of user $i$ on movie $j$ can be predicted as: $\widehat{R_{ij}}\approx E\Big[R_{ij}\Big|U_i^T{V_j},\sigma^2\Big]=U_i^T{V_j}=U_i^T\Big(\mu_j + \varepsilon_j\Big).$

\section{Ti-PMF Model}
\label{sec4}

A probabilistic neural framework is proposed in this paper to generate the descriptive documents from images. The mean idea behind the proposed model, named trailer-inception probabilistic matrix factorization (Ti-PMF), is to convert the images extracted from the movie trailers into the corresponding description texts, which will be used in the context-aware recommender system. It is possible to obtain the precise descriptive sentences of the corresponding images by directly maximizing the likelihood $p(S|I)$ of generating a target sequence of words $S$. The above description can be summarized as follows:
\begin{equation}
\theta^* = arg \max_{\theta} \sum\limits_{(I,S)} \log p(S|I;\theta),\label{eq20}
\end{equation}
where $\theta$ is the parameters of the proposed model, $I$ is a set of images, and $S$ is the set of correct transcriptions of $I$. Since $S$ represents a sentence with an unfixed length, it will be appropriate to use the chain rule to model the joint probability over $S_0$,~$\cdots$,~$S_N$. Specifically, the textual information before time $t$ and the input image information can be used to predict the textual information at $t$:
\begin{equation}
\log p(S|I) = \sum\limits _{t=0}^N \log p (S_t|I,S_0,\cdots,S_{t-1}).\label{eq21}
\end{equation}

The LSTM model can be trained to predict every word in the sentence with the prerequisite that the images and the preceding words are known, and the probability of correct prediction is defined by $p(S_t|I,S_0,\cdots,S_{t-1})$. Specifically, as for each word in a sentence, LSTM share the same parameters in all blocks. The output $m_{t-1}$ at time $t-1$ will be fed to the LSTM at time $t$. The architecture of the proposed Ti-PMF model with GoogLeNet is shown in Fig.~\ref{figure8}. Note that in the proposed Ti-PMF model, the GoogLeNet can also be replaced with VGG, the performance comparison is shown in the next section. In the Ti-PMF model, as shown in Fig.~\ref{figure8}, the NIC model combines the unrolling LSTM and GoogLeNet. Then the RCNN model is utilized to extract features of the textual information generated by NIC. After that, the RConvMF algorithm in Sect.~\ref{sec3.2} can be used to predict the ratings. The procedure of the NIC model is represented as follows in Eqs.~(\ref{eq22}), (\ref{eq23}) and (\ref{eq24}):
\begin{equation}
x_{-1} = CNN(I),\label{eq22}
\end{equation}
\begin{equation}
x_t = W_e S_t, \label{eq23}
\end{equation}
\begin{equation}
p_{t+1} = LSTM(x_t), \label{eq24}
\end{equation}
where $t\in \lbrace{0, \cdots, N-1}\rbrace$, $x_{-1}$ represents that the visual feature of image $I$, which is input into the NIC model at $t=-1$, $S_t$ represents the textual information encoded by one-hot, $W_e$ is the word2vec~\cite{T.Mikolov} model transition matrix to convert $S_t$ into a dense numeric matrix. Finally, the textual features are input into the PMF model to get an accurate rating prediction. The negative log-likelihood of the correct word at each step can be expressed as:
\begin{equation}
L(I,S) = -\sum\limits_{t=1}^N \log p_t (S_t). \label{eq25}
\end{equation}

The above loss function can be minimized over all the parameters in the LSTM model, the top layer of the image embedder CNN and the word embeddings $W_e$.

\section{Experiments}
\label{sec5}

\subsection{Datasets}
\begin{figure}[htbp]
  \small
  \centering
  \includegraphics[width=3.8in]{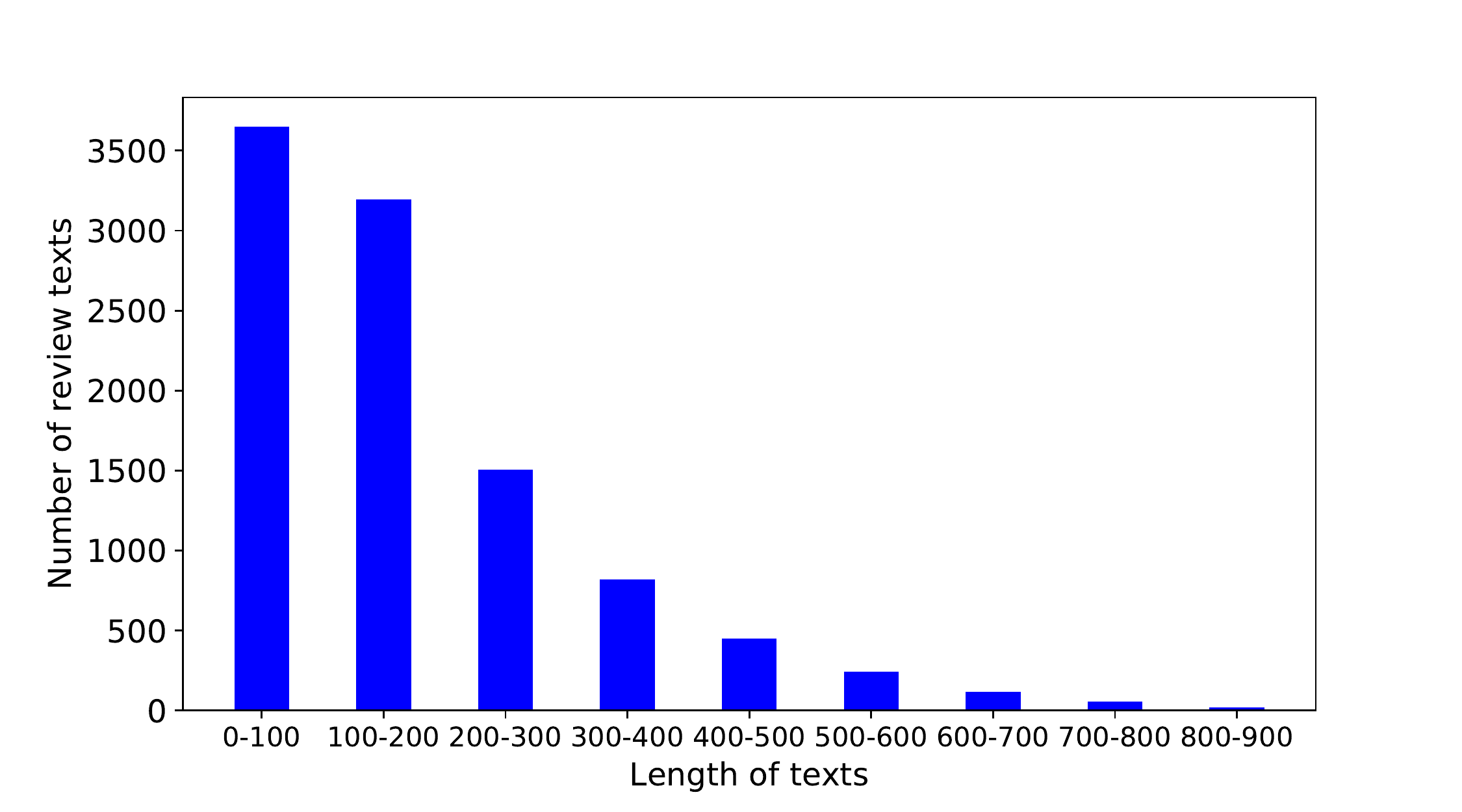}
  \caption{Data statistics of reviews in ML-10M dataset.}
  \label{figure12}
\end{figure}

The datasets used in this paper are divided into the following two parts.
\begin{itemize}
\item \textbf{Model pre-training}: To avoid overfitting, the GoogLeNet, LSTM, and RCNN models need to be pre-trained initially. Specifically, the ImageNet dataset is utilized to pre-train the GoogLeNet and VGG models, which are used for comparative experiments, and the datasets consisting of images and English sentences describing these images (such as MSCOCO, Flickr8k, and Flickr30k) are also used. The statistics of the datasets are shown in TABLE~\ref{table1}.
\item \textbf{Recommender system}: The main objective of the recommender system is to predict the target users' ratings of unknown movies. We validate the proposed Ti-PMF model on three different real-world datasets, including MovieLens-1m (marked as ML-1M), MovieLens-10m (marked as ML-10M), and Amazon Instant Video (marked as AIV). As shown in Fig.~\ref{figure12}, we perform simple sentence length statistics about ML-10M dataset on the length of text sentences. Specifically, the main statistics of the three datasets are shown in TABLE~\ref{table2}.
\end{itemize}

\begin{table}
\caption{Data statistics of pre-trained datasets.}\label{table1}
\begin{spacing}{1.4}
\centering
\begin{tabular}{|c|c|c|c|}
\hline
{Datasets}&Train&Test&Valid\\
\hline
Flickr8k&6000&1000&1000\\
\hline
Flickr30k&28000&1000&1000\\
\hline
MSCOCO&82783&40775&40504\\
\hline
\end{tabular}
\end{spacing}
\end{table}

\begin{table}
\caption{Data statistics on three real-world datasets.}\label{table2}
\begin{spacing}{1.4}
\centering
\begin{tabular}{c c c c c}
\toprule[1.5pt]
{Datasets}&Users&Movies&Ratings&Sparsity\\
\bottomrule[1.5pt]
ML-1M&6040&3883&1000209&95.73\%\\
\hline
ML-10M&71567&10681&10000054&98.69\%\\
\hline
AIV&29757&15149&135188&99.97\%\\
\toprule[1.5pt]
\end{tabular}
\end{spacing}
\end{table}
Since there is no real-world dataset of movie trailers corresponding to user ratings, we crawl the video clips of the movie ID on Youtube and IMDB. For the few early movies without trailers, we randomly select a part of the video as the trailer from the movie. Before using the NIC model to extract the image features and generate the sentence texts, we need to extract the images from the trailers. We randomly select 20 still frames as the input of the NIC model. In addition, since the item description documents are not contained in the MovieLens dataset, we get them from IMDB\footnote{http://www.imdb.com/}, Douban\footnote{https://movie.douban.com/} and Youtube \cite{S.A.-E.-H.}.

\subsection{Experiment Settings}

\subsubsection{Model Settings}
\begin{figure}
\centering
  \includegraphics[width=3.9in,height=2.6in]{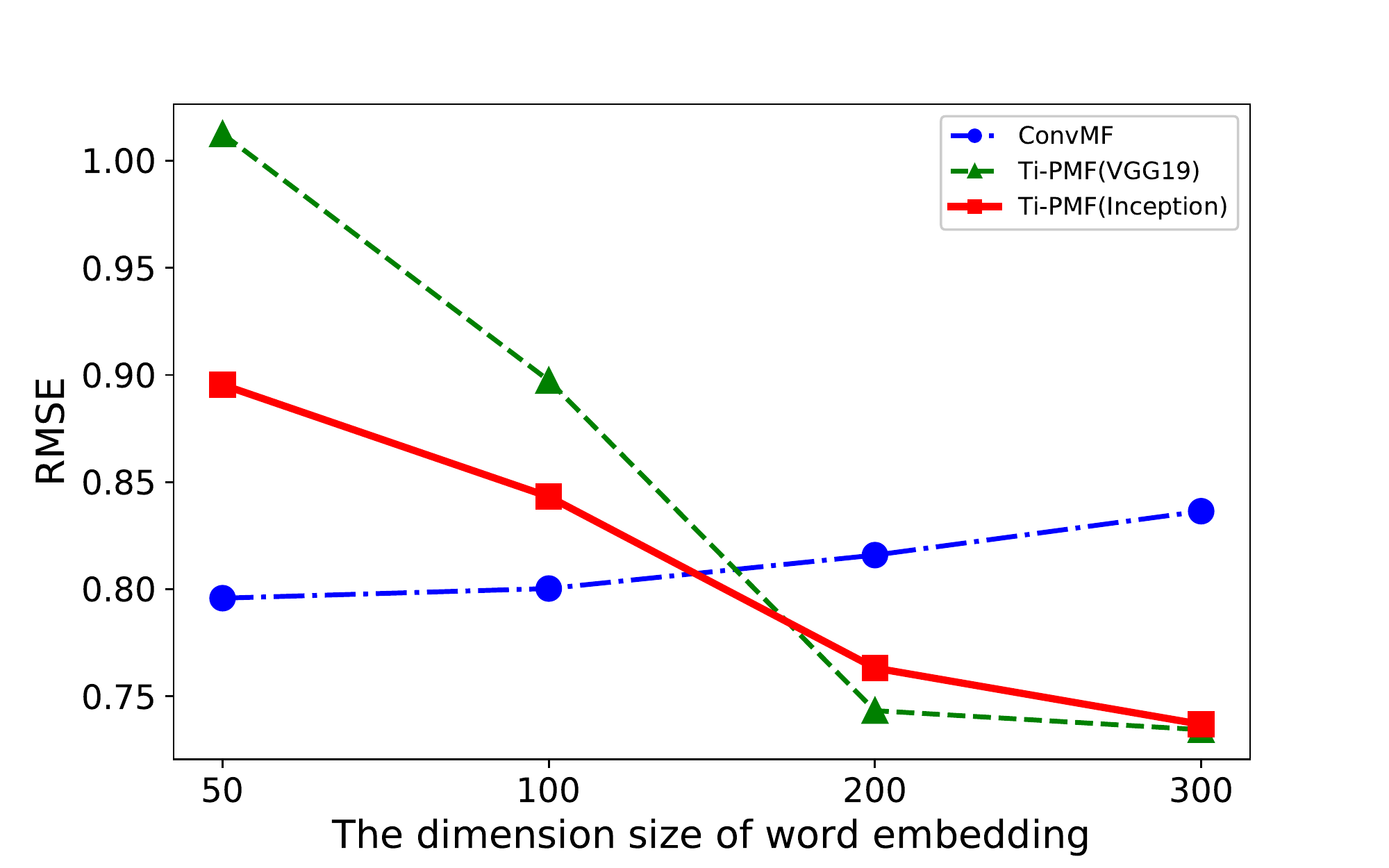}
  \caption{The effect of the dimension size of word embedding on ML-10M.}
  \label{figure11}
\end{figure}

\begin{itemize}
\item \textbf{NIC model (VGG16/19)}: Each image in the Flickr8k and Flickr30k datasets has five reference captions. Accordingly, the part of the MSCOCO dataset that exceeds five captions is deleted. All of the pre-training images in the datasets are resized into $224\times224\times3$. As the VGG network is extremely deep, the Batch-Normalization \cite{S.Ioffe} method is adopted in each layer to avoid the internal covariate shift. The dropout rate of the VGG model is set as 0.3 during the training process, and the mini-batch is set as 128.
\item \textbf{NIC model (Inception)}: The receptive field of the input image is cropped to $299\times299$ with stride 2. We put the max-pooling layer behind the first layer to reduce the parameters of the GoogLeNet, and the number of inception-a, b, c is set as 3, 5, 2, respectively. In order to enhance the effect of low-dimensional feature representation, we set the depth of the convolutional kernel to 2048 as the layers increase. At the end of the network, a max-pooling layer with $8\times8$ patch size is set to compress the features to the single-dimension deep feature vector ($1\times1\times2048$). Finally, the visual features are scaled into ($1\times1\times1000$) through a fully connected layer.
\item \textbf{NIC model (LSTM)}: We adopt the tokenization method in the NLTK library for the word labeling. The NIC model generates the texts, and the maximum raw text is set as 30. For the word embedding, the word2vec model is utilized to convert the natural language to machine language. And the word vector is initialized randomly with dimension size of 200. Fig.~\ref{figure11} shows the RMSE of Ti-PMF and ConvMF with the dimension of word embedding varying from 50 to 300. As the description texts generated by the NIC model are shorter than the comment texts, which are used as the textual information in the ConvMF model, Ti-PMF has a faster convergence speed. We will train these word latent vectors in the optimization process. Various convolutional window sizes (3, 4, 5) with 100 feature maps are utilized to obtain different contextual information.
\item \textbf{Ti-PMF}: The dimensions of user latent vector ($U$) and item latent vector ($V$) are both set as 50, and we initialize $U$ and $V$ with each element randomly selected in the range of $(0,1)$. The mini-batch size is set as 128. Thereafter, we put comprehensive features into the projection layer and fix the dimension to 50. We set the precision parameter of CTR and CDL to 1 when $r_{ij}$ is observed, otherwise it is set to 0. The number of iterations is set as 15 in Ti-PMF. An average value of five repeated experiments is performed as the final result to reduce the random error.
\end{itemize}

The NIC network is trained with stochastic gradient descent. We set the batch size as 64 for 50 epochs, and the model can be achieved using RMSProp with the decay of 0.9 and $\epsilon = 1.0$. We set the learning rate as 0.045, and the dropout rate as 0.2 to avoid over-fitting. The Ti-PMF network is trained with the coordinate descent method utilizing the Theano backend, and the dropout rate of Ti-PMF is 0.4.
\subsubsection{Evaluation Metrics}
We divide the performance evaluation of the proposed Ti-PMF model into two parts. The first part is to evaluate the precision of word n-grams. And the other part is to evaluate the rating prediction accuracy of recommender systems.
\begin{itemize}
\item \textbf{BLEU} \cite{K.Papineni}: Bilingual evaluation understudy (BLEU) is an auxiliary tool for bilingual translation quality assessment. It is a metric used to evaluate the quality of machine translation. Since manual processing is too time-consuming and laborious, the BLEU method is utilized to evaluate the quality of generated texts by machines. In our experiments, the NIC model is evaluated with four indicators (i.e., BLEU-1 to BLEU-4). Specifically, BLEU-$n$ respectively represent the value of $n$ in the $n-gram$. BLEU is a measure of the matching degree between the generated textual sequences and the texts in the ground truth.
\item \textbf{RMSE}: Before training, each dataset is randomly split into a training set \big(80\%\big), a test set \big(10\%\big) and a validation set \big(10\%\big). Then, to validate the performance of the proposed Ti-PMF model, we select the root mean squared error \big(RMSE\big) as the evaluation metric:
\begin{equation}
RMSE =\sqrt{\frac{\sum\limits_{i = 1}^m\sum\limits_{j = 1}^n\big(r_{ij}-\widehat{r_{ij}}\big)^2}{\big|N\big|}}, \label{eq26}
\end{equation}
where $\big|N\big|$ is the number of test ratings. We set the number of iterations to 30.
\end{itemize}
\begin{figure}

  \centering\includegraphics[width=3.9in,height=2.38in]{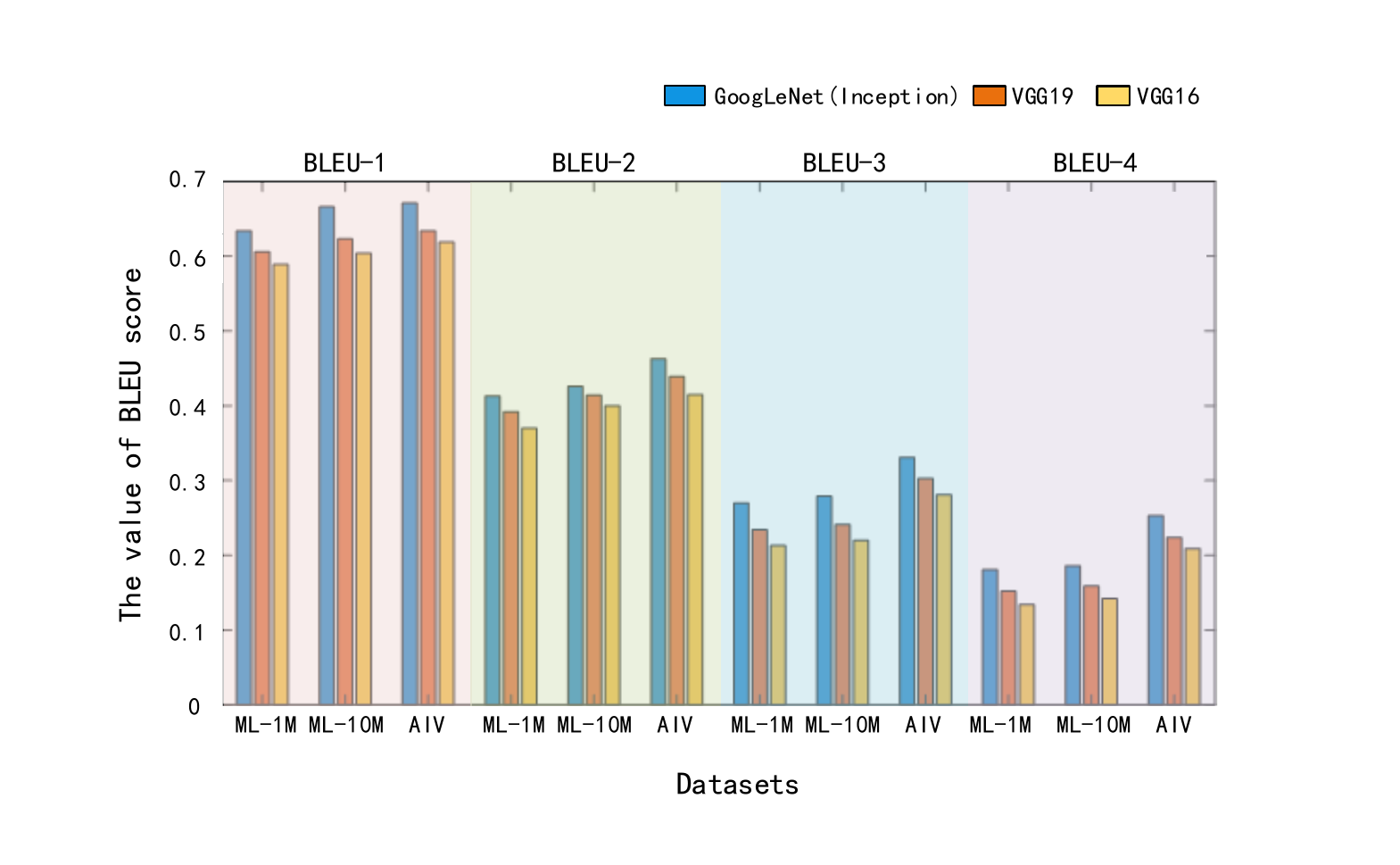}
  \caption{The BLEU score of NIC model on three different real-world datasets.}
  \label{figure9}
\end{figure}

\begin{figure}[htbp]
  \small
  \centering
  \includegraphics[width=3.8in]{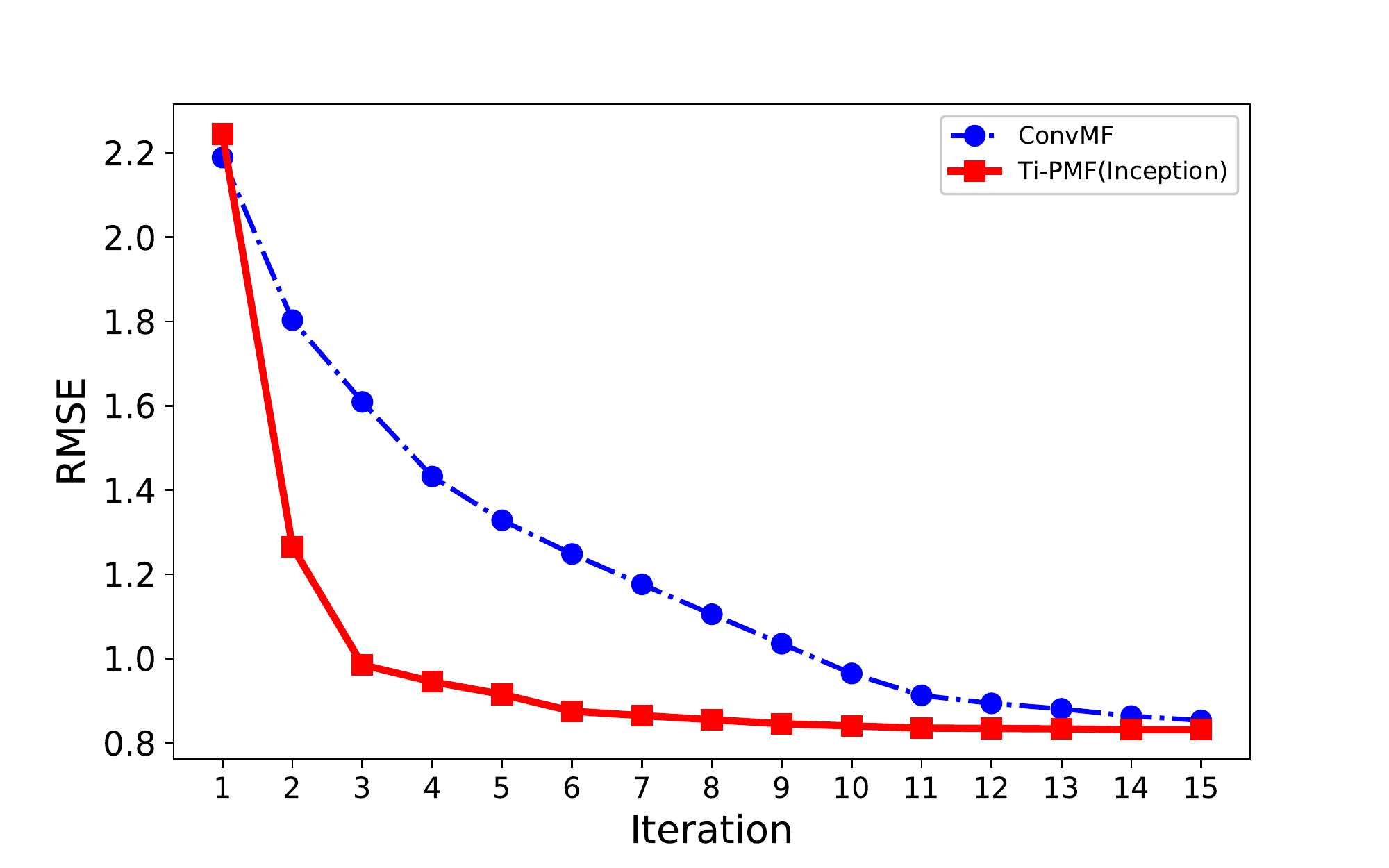}
  \caption{Comparative experiments of the iterative process between ConvMF and Ti-PMF on the ML-1m dataset.}
  \label{figure10}
\end{figure}
\subsection{Compared Schemes}

\begin{itemize}
\item \textbf{PMF} \cite{Salakhutdinov:2008}: Probabilistic matrix factorization is a basic method of rating prediction, which uses the rating scores in the form of probability.
	
\item \textbf{CTR} \cite{S.Purushotham}: Collaborative topic regression is a recommendation algorithm, which combines PMF and the latent Dirichlet allocation. Both the ratings and textual documents are used in CTR.
	
\item \textbf{CDL} \cite{H.Wang}: Collaborative deep learning is a recommendation model that improves the recommendation performance by using the stacked denoising autoencoder.
	
\item \textbf{ConvMF} \cite{D.Kim}: Convolutional matrix factorization is a context-aware recommendation model, which combines CNN and PMF seamlessly to improve the rating prediction accuracy.

\end{itemize}

\begin{figure}[htbp]
  \small
  \centering
  \includegraphics[width=3.8in]{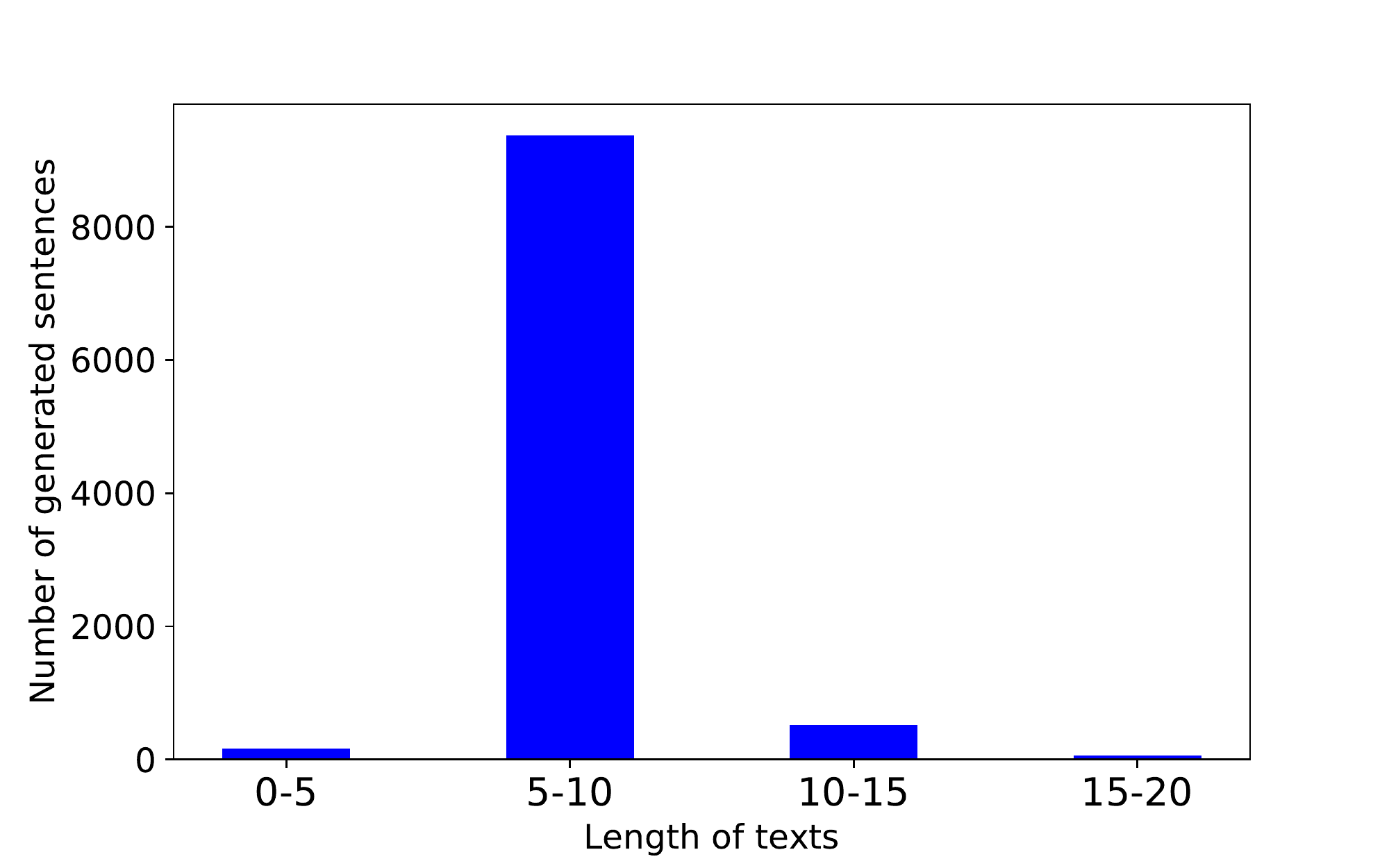}
  \caption{Data statistics of descriptive text generated by NIC model in ML-10M dataset.}
  \label{figure13}
\end{figure}
\subsection{Results and Discussion}
\subsubsection{\textbf{Comparison of VGG and Inception in NIC Model}}

In the NIC model, comparative experiments are implemented by using GoogLeNet and VGG models to connect LSTM, respectively. VGG improves the performance of CNN from the perspective of changing the network depth, while GoogLeNet improves the performance by expanding the network width. In the final result, VGG performs a bit better than GoogLeNet.  However, VGG requires more calculations during the training process. Therefore, in the application, the GoogLeNet is used to predict the ratings of the recommender systems. The experimental results compared with VGG are shown in Fig.~\ref{figure9}. BLEU-1 to BLEU-4 represent the accuracy of the generated textual information under $1-gram$ to $4-gram$ evaluation metrics. Fig.~\ref{figure9} shows that GoogLeNet achieves better performance in generating the sentences by the NIC model. Moreover, in the NIC model, GoogLeNet is superior to VGG in terms of convergence speed. The superposition of multiple convolutional kernels ($5\times5$, $3\times3$, $1\times1$) makes GoogLeNet obtain better performance than the VGG with the single-size convolutional kernel ($3\times3$).

\begin{table}
\caption{RMSE of overall test sets.}\label{table3}
\begin{spacing}{1}
\centering
\begin{tabular}{|p{2.8cm}<{\centering}|p{1.4cm}<{\centering}|p{1.4cm}<{\centering}|p{1.4cm}<{\centering}|}
\hline
\multicolumn{4}{|c|}{Dataset}\\
\hline
Model&ML-1M&ML-10M&AIV\\
\hline
PMF&0.8971&0.8311&1.4118\\
\hline
CTR&0.8969&0.8275&1.4594\\
\hline
CDL&0.8879&0.8186&1.3594\\
\hline
ConvMF&0.8531&0.7958&1.1337\\
\hline
Ti-PMF (VGG19)&\textbf{0.8120}&\textbf{0.7344}&\textbf{1.0125}\\
\hline
Ti-PMF (Inception)&0.8310&0.7369&1.0160\\
\hline
Improved&4.82\%&7.40\%&10.38\%\\
\hline

\end{tabular}
\end{spacing}

\end{table}
\subsubsection{\textbf{Impact of NIC in Ti-PMF Model}}
Inception model in Ti-PMF is indeed faster than ConvMF model as shown in Fig.~\ref{figure10}. Therefore, it can be considered that the textual information of texts generated by the NIC model is short and concise. As shown in Fig.~\ref{figure13}, for the statistics of image description texts generated by the NIC model, $98\%$ of the image description texts are with a length shorter than 10, which is much smaller than that of the users' comment texts. The generated textual information is finally input into the RConvMF model.

\subsubsection{\textbf{Overall Performance}}
TABLE~\ref{table3} shows the overall rating prediction performance of the proposed Ti-PMF and other schemes. Compared with PMF, CTR and CDL, Ti-PMF achieves significant performance improvement. Compared with ConvMF, the improvements of Ti-PMF are 4.82\%, 7.40\%, and 10.38\% on ML-1M, ML-10M and AIV datasets, respectively. The significant improvement of rating prediction performance is due to the model Ti-PMF combining RCNN and PMF to process the concise text generated by the visual information of movie trailers to extract richer contextual information, which verifies the effectiveness of the model. It is worth noting that in Ti-PMF, the training time of VGG is three times that of Inception, and better experimental results than Inception are obtained on three datasets. Moreover, the improvement of Ti-PMF on dataset AIV is much more significant than that on datasets ML-1M and ML-10M. The results in TABLE~\ref{table3} show that Ti-PMF achieves a more remarkable improvement when the dataset is with a higher sparsity, indicating that Ti-PMF can effectively alleviate the problem of data sparsity in recommender systems.

\section{Conclusions and Future Work}
\label{sec6}
In this paper, we propose a trailer inception probabilistic matrix factorization model called Ti-PMF. The proposed Ti-PMF model combines the neural image caption, recurrent convolutional neural network, and probabilistic matrix factorization model as the rating prediction model of recommender systems. We implement the proposed Ti-PMF model and conduct extensive experiments on three real-world datasets to illustrate that the proposed Ti-PMF model outperforms the existing ones.

In future work, we will take into account the user attributes (such as gender, age, and occupations) to promote the performance of rating prediction. Furthermore, we also consider using the RNN model to extract the video features to improve the recommendation performance.

\section*{ACKNOWLEDGEMENT}
This work was supported in part by NSFC grants (No.61772551), the Open Project of Minjiang University (No.MJUKF-JK202003) and the Major Scientific and Technological Projects
of CNPC under Grant ZD2019-183-003.

\bibliographystyle{elsarticle-num}
\bibliography{Ti-PMF}



\end{document}